\documentclass[11pt]{article}

\usepackage[final]{acl}

\usepackage{times}
\usepackage{latexsym}

\usepackage[T1]{fontenc}

\usepackage[utf8]{inputenc}

\usepackage{microtype}

\usepackage{inconsolata}

\usepackage{graphicx}
\usepackage[table]{xcolor}
\usepackage{booktabs}
\usepackage{caption}
\usepackage{tabularx}
\usepackage{ragged2e}
\usepackage{array}
\usepackage{subcaption}
\usepackage{amsmath}
\usepackage{algorithm}
\usepackage{algpseudocode}
\usepackage{hyperref}

\title{SAM-NER: Semantic Archetype Mediation for Zero-Shot \\ Named Entity Recognition}

\author{
 \textbf{Ruichu Cai\textsuperscript{1,2}},
 \textbf{Juntao Gan\textsuperscript{1}},
 \textbf{Miao Mai\textsuperscript{3}},
 \textbf{Zhifeng Hao\textsuperscript{1,4}},
 \textbf{Boyan Xu\textsuperscript{1}}\thanks{Corresponding author.}
 \\
 \textsuperscript{1}School of Computer Science, Guangdong University of Technology \\
 \textsuperscript{2}Peng Cheng Laboratory \quad
 \textsuperscript{3}Nanfang Media Group(Nanfang Daily) \\
 \textsuperscript{4}College of Mathematics and Computer, Shantou University \\
 \{cairuichu, yinghuo.gan\}@gmail.com \quad maim@nfmedia.com \\
 haozhifeng@stu.edu.cn \quad hpakyim@gmail.com
}

\begin{document}
\maketitle
\begin{abstract}
Zero-shot Named Entity Recognition (ZS-NER) remains brittle under domain and schema shifts, where unseen label definitions often misalign with a large language model's (LLM's) intrinsic semantic organization. As a result, directly mapping entity mentions to fine-grained target labels can induce systematic semantic drift, especially when target schemas are novel or semantically overlapping. We propose \textbf{SAM-NER}, a three-stage framework based on \emph{Semantic Archetype Mediation} that stabilizes cross-domain transfer through an intermediate, domain-invariant archetype space. SAM-NER: (i) performs \emph{Entity Discovery} via cooperative extraction and consensus-based denoising to obtain high-coverage, high-fidelity entity spans; (ii) conducts \emph{Abstract Mediation} by projecting entities into a compact set of universal semantic archetypes distilled from high-level ontological abstractions; and (iii) applies \emph{Semantic Calibration} to resolve archetype-level predictions into target-domain types through constrained, definition-aligned inference with a frozen LLM. Experiments on the CrossNER benchmark show that SAM-NER consistently outperforms strong prior ZS-NER baselines in cross-domain settings. Our implementation will be open-sourced at \url{https://github.com/DMIRLAB-Group/SAM-NER}.
\end{abstract}

\section{Introduction}

Zero-shot Named Entity Recognition (ZS-NER) aims to identify and type entities in unseen domains or under novel label taxonomies without target-domain supervision \cite{chatIE, Self-Improving}. With their strong general language understanding and broad world knowledge, large language models (LLMs) have become the dominant paradigm for tackling ZS-NER, enabling flexible reasoning over natural language descriptions of entity types and extraction rules. Most existing LLM-based ZS-NER approaches can be broadly categorized into two paradigms: (\romannumeral 1) instruction-based structured extraction via natural language constraints, and (\romannumeral 2) retrieval-augmented inference (RAG) that incorporates external evidence during generation or verification.

Instruction-based methods reformulate entity specifications and task constraints into prompts that guide LLMs to perform span extraction and type assignment. For example, \citet{GOLLIE} maps structured task specifications into natural-language annotation instructions, allowing LLMs to follow explicit extraction principles across diverse domains. In contrast, retrieval-augmented approaches enhance inference by incorporating exogenous knowledge sources; for instance, \citet{verifiner} retrieves background evidence to support post-hoc verification and error correction, aiming to reduce hallucinations and improve reliability. Both paradigms have demonstrated promising performance under zero-shot and low-resource settings, highlighting the potential of LLMs as a unifying backbone for NER.

\begin{figure}[t]
  \includegraphics[width=1 \columnwidth]{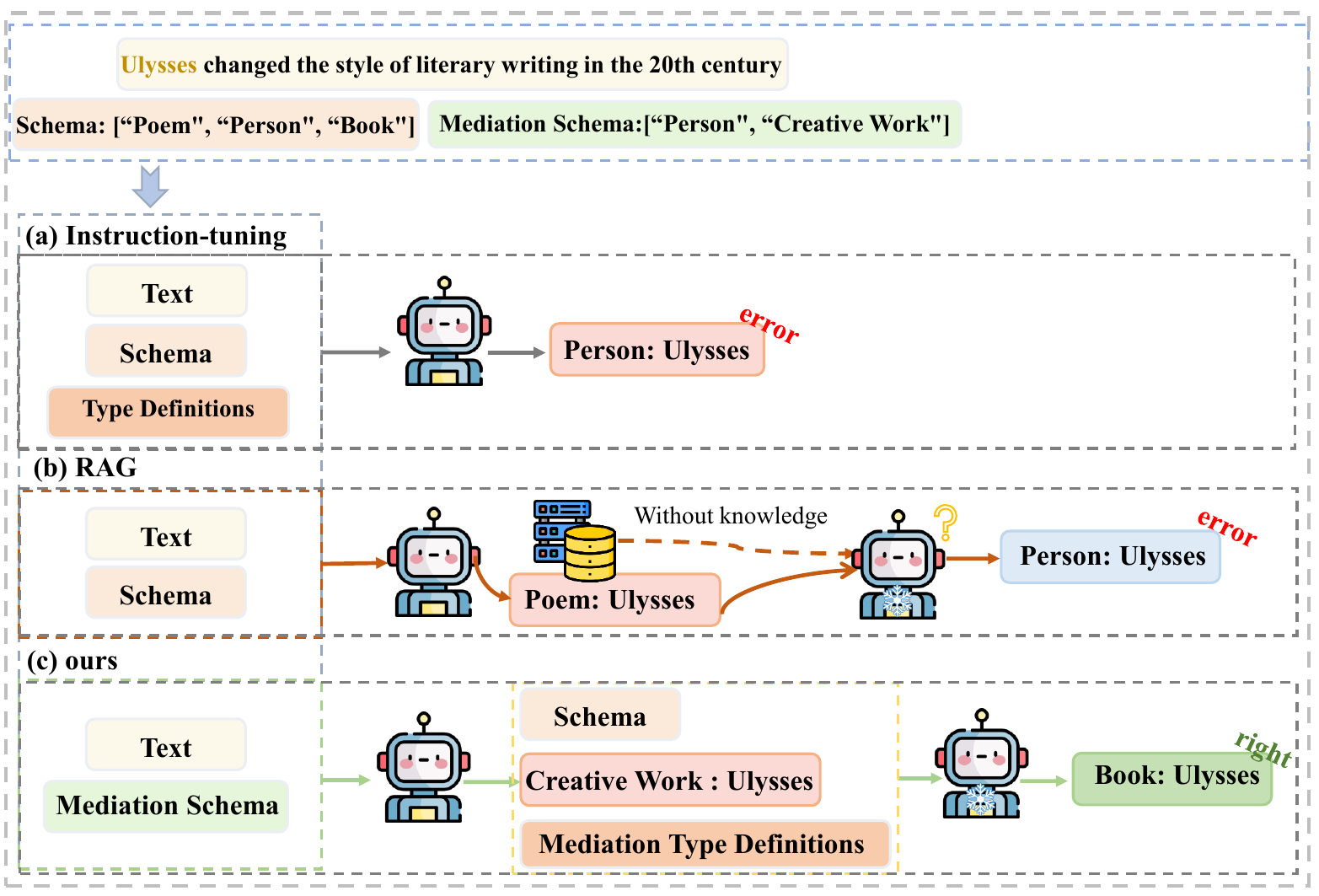}
  \caption{Comparison of different ZS-NER methods}
  \label{fig:motivation}
\end{figure}

However, existing LLM-based ZS-NER methods exhibit fundamental limitations when applied to cross-domain settings with heterogeneous label semantics. Instruction-based approaches implicitly assume that target-domain label definitions align well with the model’s intrinsic semantic organization—an assumption that often fails when fine-grained, domain-specific schemas introduce systematic semantic drift. Meanwhile, retrieval-augmented methods are inherently constrained by the availability, coverage, and reliability of external knowledge sources, which are frequently sparse or incomplete in specialized vertical domains. As illustrated in Figure~\ref{fig:motivation}(a) and (b), both forms of mismatch—semantic misalignment between model representations and target definitions, and insufficiencies in external knowledge—lead to characteristic failure modes in zero-shot extraction, resulting in distorted or erroneous entity predictions. These observations motivate the need for an intermediate semantic abstraction that can stabilize cross-domain transfer without relying on domain-specific supervision or external knowledge.

To address these challenges, we propose \textbf{SAM-NER}, a zero-shot NER framework based on the principle of \emph{Semantic Archetype Mediation}. Our key insight is that while target-domain label taxonomies are often volatile and domain-specific, the underlying semantic archetypes instantiated by entities remain largely invariant across domains. Instead of forcing LLMs to directly map entity mentions to narrow and unseen target labels, SAM-NER introduces an intermediate mediation layer that stabilizes semantic reasoning by decoupling semantic understanding from schema-specific label definitions. SAM-NER operates as a three-stage pipeline. 
(\romannumeral 1) \emph{Entity Discovery via Cooperative Extraction} identifies a high-recall yet high-fidelity set of candidate entity spans through dual-source extraction and consensus-based denoising; 
(\romannumeral 2) \emph{Abstract Mediation via Semantic Archetypes} projects these candidates into a universal archetype space, establishing stable and domain-invariant semantic anchors; and 
(\romannumeral 3) \emph{Definition-Guided Semantic Calibration} resolves archetype-level predictions into fine-grained target-domain types through constrained, definition-aligned inference with a frozen LLM. As illustrated in Figure~\ref{fig:motivation}(c), this mediation-driven process enables reliable discrimination between semantically adjacent target types even under severe schema shift.

The main contributions of this work are summarized as follows:
\begin{itemize}
    \item We introduce \emph{Semantic Archetype Mediation}, a new paradigm for zero-shot NER that stabilizes cross-domain generalization by decoupling semantic understanding from volatile label definitions.
    \item We propose a cooperative entity discovery framework that combines a precision-oriented anchor extractor with a recall-oriented explorer extractor, and introduce consensus-based denoising to reconcile their complementary error profiles.
    \item We develop a definition-guided semantic calibration mechanism that grounds archetype-level predictions into target-domain labels through constrained, definition-aligned reasoning, eliminating reliance on external knowledge bases.
    \item Extensive experiments on CrossNER demonstrate that SAM-NER achieves state-of-the-art performance, validating semantic archetype mediation as a more robust alternative to direct label mapping for out-of-domain generalization.
\end{itemize}

\begin{figure*}[t]
  \includegraphics[width=1\linewidth]{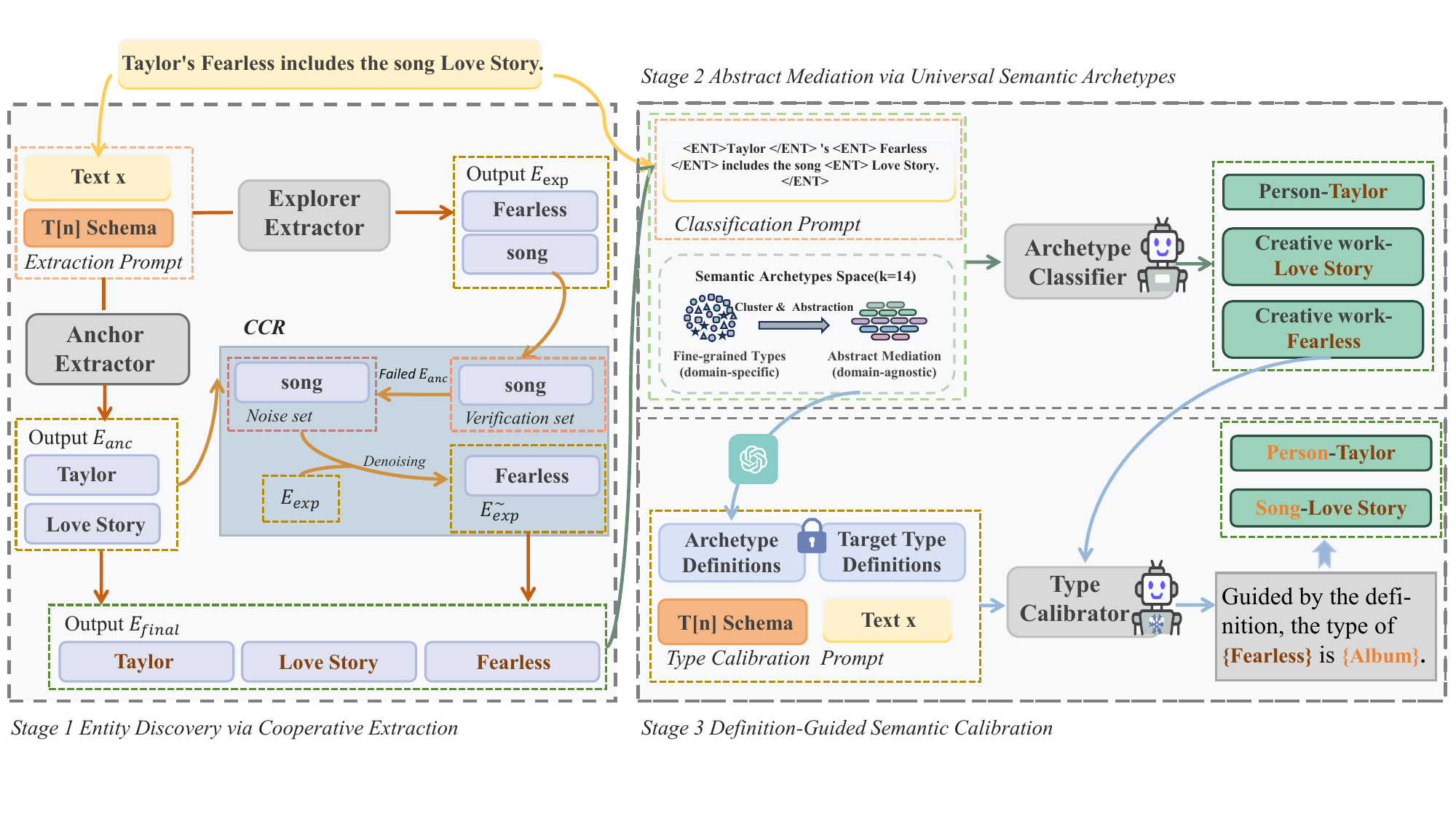}
  \caption {\textbf{The overall framework}. In Stage 1, extracted entities are used to annotate sentences. In Stage 2, archetype mediation are assigned to the marked entities. In Stage 3, the mediation types are refined into target domain types based on type definitions.}
  \label{fig:framework}
\end{figure*}

\section{Related Work}

\paragraph{Generative Models for NER} Instruction tuning is a paradigm that was widely adopted in NER during the early stages of large language model development. Prior work injects task priors and entity type semantics into LLMs through natural language instructions, such as annotation guidelines, extraction rules, and type descriptions, which has been shown to effectively improve entity boundary detection and type discrimination, particularly in zero-shot and low resource settings \cite{instructuie, universalNER, schemaTable, advancingBiomedicine}.

To further enhance the stability of instruction representations, recent studies introduce more structured formulations. Approaches such as Code4UIE \cite{Code4UIE} and KnowCoder \cite{knowcoder} encode extraction schemas and type constraints as executable code or structured class hierarchies, explicitly modeling semantic relations among entity types while retaining the core instruction following paradigm.

In addition, with the further development of large language models, agent-based methods have emerged as a significant direction in NER research. This approach decomposes NER into multiple interactive agents and enhances robustness in zero-shot or low-resource scenarios through structured interactions, tool usage, or iterative improvements. Key mechanisms include: self-iteration/reflection, which involves reviewing and correcting predictions or pseudo-labels to refine entity boundaries and types \cite{CMAS, OEMA}; and networked retrieval, which utilizes an independent Knowledge Retrieval Agent to obtain external knowledge from Wikipedia to enrich contextual information \cite{KDR-Agent}. 

Despite their effectiveness, instruction tuning methods generally assume that LLM internal semantic representations are well aligned with target domain entity type semantics. This assumption often breaks down in out-of-domain or novel type scenarios, where semantic misalignment limits generalization performance. In contrast, agent-based methods leverage multi-step collaboration among different agents to enhance the accuracy of reasoning, with performance improvements primarily stemming from external coordination and multi-step collaboration.

\paragraph{Entity Type Descriptions without External Knowledge} In knowledge intensive Named Entity Recognition scenarios, retrieval-augmented generation (RAG) has been widely adopted to enhance entity recognition and type discrimination. Such approaches typically assist LLMs by retrieving similar annotated examples from demonstration pools or by incorporating entity-related background knowledge from external knowledge bases during inference \cite{IF-WRANER, RUIE, IRRA, advancingBiomedicine}. Although these methods achieve substantial performance gains across multiple benchmarks, their effectiveness heavily depends on the quality and coverage of external knowledge resources, limiting their scalability to data-scarce domains.

In the absence of external knowledge, recent studies have explored leveraging natural language descriptions—such as entity type definitions—to provide semantic constraints for NER \cite{GOLLIE, showless, openbioner, zeroner}. Instead of treating entity types as discrete symbolic labels, these methods convert them into human-readable forms, including type definitions, annotation guidelines, or labeling instructions, and incorporate them as soft constraints during extraction and classification. This formulation enables LLMs to perform entity typing based on semantic understanding, even without explicit domain knowledge.

Nevertheless, these approaches implicitly rely on a stable alignment between target type semantics and the model’s intrinsic semantic space. Under domain shift or unseen label settings, directly reasoning over target type definitions can introduce semantic mismatch, leading to type ambiguity or unstable predictions. This limitation motivates further investigation into how type semantics can be leveraged more robustly for out-of-domain NER without external knowledge.

\section{Methodology}
\textbf{SAM-NER} is a three-stage framework designed to mitigate semantic drift in zero-shot Named Entity Recognition through \emph{semantic archetype mediation}. As illustrated in Figure~ \ref{fig:framework}, the framework follows a progressive mediation pipeline that transitions from entity span discovery to abstract semantic anchoring, and finally to definition-grounded target typing.
Specifically, 
(i) the \textbf{Entity Discovery} stage via Cooperative Extraction identifies a high-recall yet high-fidelity set of candidate entity spans through dual-source extraction and consensus-based denoising; 
(ii) the \textbf{Abstract Mediation} stage via Universal Semantic Archetypes projects these candidates into a universal archetype space, establishing stable and domain-invariant semantic anchors; and 
(iii) the Definition-Guided \textbf{Semantic Calibration} stage resolves archetype-level predictions into fine-grained target-domain types through constrained, definition-aligned inference.
By decoupling entity discovery, semantic mediation, and target-domain grounding, SAM-NER provides a robust and interpretable solution for zero-shot NER under heterogeneous label semantics.

\subsection{Entity Discovery via Cooperative Extraction}
This stage aims to construct a high-coverage yet high-fidelity set of \emph{type-agnostic} entity span candidates for subsequent semantic archetype mediation. We adopt a dual-source design to exploit complementary error profiles: a precision-oriented \emph{anchor extractor} provides stable corroboration signals, while a silver-supervised \emph{explorer extractor} improves recall but may introduce spurious mentions. We then perform consensus-based denoising to suppress silver-noise without sacrificing coverage.

\paragraph{Anchor Extractor.}
We instantiate the anchor extractor \(Extractor_{\text{anc}}\) as a precision-oriented instruction-tuned model. Specifically, we use Llama3-8B-Instruct fine-tuned on high-quality IE instructions from IEPile \cite{iepile}, which exhibits strong boundary discrimination and stable semantic behavior under domain shift. Given an input sentence \(x_i\) and a target type set \(T_i\), the anchor extractor outputs a set of candidate entity mentions:
\begin{equation}
E_{\text{anc}}^{i} = Extractor_{\text{anc}}(x_i, T_i).
\label{eq:anchor_extractor}
\end{equation}
Each \(e \in E_{\text{anc}}^{i}\) denotes an extracted entity span in \(x_i\).

\paragraph{Explorer Extractor.}
To increase candidate coverage and capture entities that may be missed by the anchor extractor, we introduce a recall-oriented explorer extractor \(Extractor_{\text{exp}}\) trained with broad-spectrum silver supervision. We train \(Extractor_{\text{exp}}\) using Pile-NER \cite{universalNER} by converting its dialogue-style supervision into a standard instruction-following format. Given the same input \((x_i, T_i)\), the explorer extractor produces an additional candidate set:
\begin{equation}
E_{\text{exp}}^{i}
=
Extractor_{\text{exp}}(x_i, T_i).
\label{eq:explorer_extractor}
\end{equation}

In practice, \(Extractor_{\text{exp}}\) tends to over-generate low-salience, word-level mentions (e.g., "data", "system", "user"). We attribute this behavior to noise in the \textbf{silver-standard annotations} of Pile-NER, where automated labeling may spuriously mark generic functional nouns as entities, particularly under domain shift. Such spurious candidates introduce substantial variance and motivate an explicit denoising mechanism before archetype mediation.

\paragraph{Collaborative Consensus Refinement.}
We introduce Collaborative Consensus Refinement (CCR) as a consensus-based denoising mechanism to refine the explorer candidates produced by the silver-supervised extractor. CCR leverages the anchor extractor as an independent semantic validator and selectively filters noise-prone candidates in \(E_{\text{exp}}^{i}\) through cross-model consensus. The underlying intuition is that generic, non-referential spans are prone to appear in silver-supervised outputs, whereas the instruction-tuned anchor extractor provides a stronger semantic prior for validating whether a span constitutes a meaningful entity mention in context.

Concretely, we identify a subset of explorer candidates that are most prone to silver-noise artifacts and aggregate them into a verification set \(V_{\text{noise}}^{i} \subseteq E_{\text{exp}}^{i}\). A candidate \(e \in V_{\text{noise}}^{i}\) is marked as noise only if it fails to obtain independent corroboration from the anchor extractor:
\begin{equation}
D_{\text{noise}}^{i}
=
\{\, e \in V_{\text{noise}}^{i} \mid e \notin E_{\text{anc}}^{i} \,\}.
\label{eq:ccr_noise_set}
\end{equation}

The denoised explorer candidate set is then obtained as:
\begin{equation}
\tilde{E}_{\text{exp}}^{i}
=
E_{\text{exp}}^{i} \setminus D_{\text{noise}}^{i}.
\label{eq:ccr_denoised_exp}
\end{equation}

Finally, we construct the discovered entity set via a de-duplicated union:
\begin{equation}
E_{\text{final}}^{i}
=
E_{\text{anc}}^{i} \cup \tilde{E}_{\text{exp}}^{i}.
\label{eq:ccr_final_union}
\end{equation}
This consensus-driven refinement preserves the high-precision predictions of the anchor extractor while retaining the coverage benefits of the explorer extractor, yielding a high-fidelity candidate set for subsequent semantic archetype mediation.

\noindent\rule{\linewidth}{0.8pt}
\noindent \textbf{Algorithm 1: Collaborative Consensus Refinement (CCR)}  \hfill \\
\rule{\linewidth}{0.8pt}

\begin{algorithmic}[1]

\State \textbf{Input:} $E_{anc}, E_{exp}$
\State $D_{noise} \gets \emptyset,\; \tilde{E}_{\text{exp}}^{i} \gets \emptyset$

\For{ $e^i_{exp} \in E_{exp}$ }

    \State $len_e \gets len(\text{split}(e^i_{exp}))$
    
    \If{ $len_e = 1 \;\And\; e^i_{exp} \notin E_{anc}$ }

        \State $D_{noise} \gets D_{noise} \cup \{ e^i_{exp} \}$
        
    \EndIf

\EndFor
    
\For{ $e^i_{exp} \in E_{exp}$ }

    \If{ $e^i_{exp} \notin D_{noise}$ }

        \State $\tilde{E}_{\text{exp}}^{i} \gets \tilde{E}_{\text{exp}}^{i} \cup \{ e^i_{exp} \}$

    \EndIf

\EndFor

\State \textbf{Output:} $\tilde{E}_{\text{exp}}^{i}$

\end{algorithmic}
\noindent\rule{\linewidth}{0.8pt}

\subsection{Abstract Mediation via Universal Semantic Archetypes}
The core bottleneck in zero-shot Named Entity Recognition lies in the high variance and domain-specificity of target label semantics. 

To address this challenge, we introduce an intermediate \emph{Abstract Mediation} stage—\emph{Semantic Archetype Mediation}—that bridges entity discovery and target-domain calibration. Our key hypothesis is that while fine-grained entity labels vary substantially across domains and annotation schemas, they can be projected into a compact and stable semantic space consisting of broad, invariant archetypes. By performing classification within this archetype space, we decouple semantic understanding from volatile label definitions and reduce the alignment pressure between the LLM's intrinsic representations and heterogeneous domain-specific schemas.

\paragraph{Ontological Distillation and Archetype Mapping.}
We construct the mediation space by consolidating heterogeneous NER schemas from IEPile \cite{iepile}, which integrates multiple information extraction benchmarks with instance-specific type sets. Let \(D_{\text{ner}}\) denote the NER subset of IEPile. For each training instance \(i\), let \(\mathcal{T}_{\text{orig}}^{i}\) denote its original fine-grained type schema, which may differ substantially across data sources.

Guided by taxonomic design principles proposed by \citet{beyondboundaries}, we distill a universal set of 14 semantic archetypes \(\mathcal{A}\) (e.g., \textit{Person}, \textit{Medicine}) to serve as the shared mediation space for zero-shot NER. Rather than adopting an existing ontology verbatim, we construct an abstract type system explicitly tailored for semantic mediation across heterogeneous NER schemas. The choice of archetype granularity is empirically motivated and further analyzed in Section~\ref{sec:analysis}.

The resulting archetypes are designed to satisfy several desiderata: 
(i) \emph{readability}, ensuring that type names are interpretable to both humans and language models; 
(ii) \emph{unambiguity}, minimizing semantic overlap between closely related categories; 
(iii) \emph{hierarchical coherence}, aligning fine-grained types with their abstract parents; and 
(iv) \emph{flexibility}, allowing the taxonomy to accommodate diverse NER tasks and emerging entity types.

We formalize schema consolidation via a deterministic projection function:
\begin{equation}
M:\mathcal{T}_{\text{orig}} \rightarrow \mathcal{A},
\label{eq:type_projection}
\end{equation}
where \(\mathcal{T}_{\text{orig}}=\bigcup_i \mathcal{T}_{\text{orig}}^{i}\) denotes the union of fine-grained types observed in \(D_{\text{ner}}\). For each instance \(i\), the corresponding abstract schema is obtained by projecting its original schema:
\begin{equation}
\mathcal{A}^{i} = M(\mathcal{T}_{\text{orig}}^{i}) \subseteq \mathcal{A}.
\label{eq:instance_schema_projection}
\end{equation}

The projection function \(M\) is constructed through a data-informed, principle-guided schema design process. Semantically related fine-grained labels are grouped under shared abstract parents according to their conceptual scope and hierarchical relations, while enforcing the desiderata above. Importantly, \(M\) is deterministic and fixed throughout training and inference. For reproducibility, the complete type-to-archetype mapping is provided in Appendix~\ref{app:appendixC}.

Applying \(M\) to the original annotations resolves ontological overlaps and homogenizes the label space. To preserve the relationship between each entity mention and its sentential context, we employ an entity-aware formatting strategy that marks mentions with \texttt{<ENT>} and \texttt{</ENT>} tags. The resulting abstract training set is defined as:
\begin{equation}
D_{\text{abs}}=\{(S_{\text{tag}}^{i},\, \mathcal{A}^{i},\, Y_{\text{abs}}^{i})\}_{i=1}^{|D_{\text{ner}}|},
\label{eq:abstract_training_set}
\end{equation}
where \(S_{\text{tag}}^{i}\) is the tagged sentence and \(Y_{\text{abs}}^{i}\) denotes the archetype-level annotations obtained by projecting the original labels through \(M\). This design allows the classifier to respect instance-specific schema constraints during training, while learning a unified and domain-invariant abstraction space.

\paragraph{Abstract Archetype Classifier.}
We train an abstract archetype classifier \(Classifier_{\text{abs}}\) on \(D_{\text{abs}}\) to serve as the mediation module. During training, the classifier is conditioned on the instance-specific abstract schema \(\mathcal{A}^{i}\), reflecting the multi-source setting where the candidate type set varies across examples. During inference, since the input consists only of entity-marked sentences without schema constraints, the classifier predicts over the full archetype space \(\mathcal{A}\):
\begin{equation}
y_{\text{abs}}^{i}
=
Classifier_{\text{abs}}(x_{\text{tag}}^{i}, \mathcal{A}),
\label{eq:abstract_inference}
\end{equation}
where \(y_{\text{abs}}^{i}\) represents the predicted archetype assignments for entities in the sentence. The collection \(Y_{\text{abs}}=\{y_{\text{abs}}^{i}\}_{i=1}^{n}\) provides stable semantic anchors for the subsequent definition-guided calibration stage, enabling target-domain typing to be grounded in a pre-aligned and domain-invariant semantic framework.

\subsection{Definition-Guided Semantic Calibration}
To bridge the final gap between the abstract archetype space \(\mathcal{A}\) and domain-specific taxonomies, we introduce a \emph{Definition-Guided Semantic Calibration} stage. Given the archetype predictions \(Y_{\text{abs}}\) produced by the semantic archetype mediation stage, this module resolves each archetype assignment into a fine-grained target-domain label through \emph{constrained type inference}. Unlike approaches that directly prompt LLMs to predict target labels, we leverage archetypes as semantic anchors and explicitly condition target-type reasoning on the predicted \(y_{\text{abs}}^{i}\), thereby narrowing the inference space and mitigating semantic drift in zero-shot transfer.

\paragraph{Axiomatic Definition Construction.}
To establish clear and stable semantic boundaries for the archetype space, we associate each abstract archetype \(a \in \mathcal{A}\) with a canonical semantic definition. Let
\begin{equation}
\mathcal{D}_{\text{abs}} = \{ d_a \mid a \in \mathcal{A} \}
\label{eq:abs_definitions}
\end{equation}
denote the set of abstract type definitions, where each \(d_a\) characterizes the essential semantic scope of archetype \(a\). These definitions are synthesized via prompt-based distillation using a large language model \cite{chatgpt2025} and are designed to be stylistically consistent and semantically exclusive. As a result, they serve as explicit semantic constraints that delineate the abstract mediation space without introducing domain-specific leakage.

For a given target domain, let \(\mathcal{T}_{\text{tgt}}\) denote its entity type set. We associate each target type \(t \in \mathcal{T}_{\text{tgt}}\) with a refined natural language definition and denote the resulting definition set as:
\begin{equation}
\mathcal{D}_{\text{tgt}} = \{ d_t \mid t \in \mathcal{T}_{\text{tgt}} \}.
\label{eq:tgt_definitions}
\end{equation}
We apply minimal linguistic normalization to the original benchmark descriptions (e.g., CrossNER \cite{crossner}) to remove syntactic redundancies and underspecified semantics while preserving their original intent. These refined target definitions constitute the axiomatic constraints for definition-guided calibration.

\paragraph{Constrained Definition-Aligned Inference.}
We employ a frozen large language model as a calibration operator, denoted by \(Calibrator\), to perform definition-aligned type inference. For each extracted entity instance \(i\), we are given its context sentence \(S^{i}\), its predicted abstract archetype \(y_{\text{abs}}^{i} \in \mathcal{A}\), and the target-domain type definitions \(\mathcal{D}_{\text{tgt}}\). The calibration objective is to resolve the most compatible target type under the constraint imposed by the archetype prior:
\begin{equation}
y_{\text{tgt}}^{i}
=
\arg\max_{t \in \mathcal{T}_{\text{tgt}}}
\; \text{Align}\big(d_{y_{\text{abs}}^{i}}, d_t \mid S^{i}\big),
\label{eq:calibration_objective}
\end{equation}
where \(d_{y_{\text{abs}}^{i}} \in \mathcal{D}_{\text{abs}}\) is the definition of the predicted archetype, and \(\text{Align}(\cdot)\) denotes a definition-alignment scoring function implicitly realized by the frozen LLM. This function evaluates the semantic entailment and compatibility between the archetype prior and candidate target definitions within the given context.

By enforcing this abstract-to-target alignment constraint, the calibrator avoids unconstrained label reasoning and instead grounds target-type selection in a pre-stabilized semantic framework. This dual-level alignment enables robust discrimination between semantically adjacent target types, even when they are unseen during training.

Finally, the calibrated predictions across all extracted entities are aggregated as:
\begin{equation}
Y_{\text{tgt}} = \{ y_{\text{tgt}}^{i} \}_{i=1}^{N},
\label{eq:final_predictions}
\end{equation}
where \(N\) is the number of extracted entity instances. This calibrated entity set constitutes the final output of \textbf{SAM-NER}.

\begin{table*}[t]
\centering
\small
\setlength{\tabcolsep}{6pt}
\begin{tabular}{l|rr|ccccc|c}
\toprule
\textbf{Method} & \textbf{Params} & \textbf{Backbone} & 
\textbf{AI} & \textbf{Literature} & \textbf{Music} & \textbf{Politics} & \textbf{Science} & \textbf{Avg.} \\
\midrule
InstructUIE (2023)     & 11B & Flan-T5    & 49.0 & 42.7 & 53.2 & 48.1 & 49.2 & 48.4 \\
UniNER (2023)   & 13B & LLaMA      & 54.2 & 60.9 & 64.5 & 61.4 & 63.5 & 60.9 \\
IEPile-Llama3-8B (2024)           & 8B & Llama3   &  50.2  &  43.3  &  53.7  &  57.0  &  50.4  & 50.9 \\
GoLLIE (2024)          & 34B  & CodeLLaMA  & \underline{61.6} & 59.1 & 68.4 & 60.2 & 56.3 & 61.1 \\
KnowCoder (2024)       & 7B  & LLaMA2     & 60.3 & 61.1 & \underline{70.0} & \underline{72.2} & 59.1 & 64.5 \\
GLiNER-Large (2024) & 0.3B  & DeBERTa-v3 & 57.2 & \underline{64.4} & 69.6 & \textbf{72.6} & 62.6 & 65.3\\
IRRA-Guidelines (2025) & 8B  & LLaMA3     & 53.2 & 57.7 & 64.7 & 66.2 & 64.1 & 61.2 \\
GUIDEX (2025)          & 8B  & LLaMA3.1   & \textbf{62.4} & 63.8 & 67.9 & 69.6 & \underline{64.6} & \underline{65.7} \\
\midrule
\textbf{SAM-NER}(Qwen2.5-7B) & 7B & Qwen2.5 & 57.9 & 64.1 & 69.3 & 66.7 & 62.1 & 64.3 \\
\textbf{SAM-NER}(Llama3-8B)                & 8B  & LLaMA3     & 58.2 & \textbf{68.7} & \textbf{71.2} & 68.2 & \textbf{65.1} & \textbf{66.3} \\
\bottomrule
\end{tabular}
\caption{The micro-F1 scores on zero-shot cross-domain setting. Except for IRRA, all baseline scores are directly taken from the results reported in the original papers under their optimal settings. We used the scores generated by IRRA under the Guidelines settings for comparison.}
\label{tab:result}
\end{table*}

\section{Experimental Settings}

\subsection{Datasets}
\paragraph{Training Sets.} In our experiments, we use Pile-NER \cite{universalNER} and IEPile \cite{iepile} as training datasets. Pile-NER covers approximately 13K entity types and contains around 240K entity instances, exhibiting a highly diverse distribution of entity types. IEPile is an instruction-tuning dataset for information extraction on a large scale that integrates 33 widely used IE benchmarks, in this work, we use only its Named Entity Recognition subset.
\paragraph{Benchmark.} We adopt CrossNER \cite{crossner} as the evaluation benchmark to assess the effectiveness of the proposed approach on out-of-domain zero-shot Named Entity Recognition. CrossNER spans multiple domains, including Artificial Intelligence, Literature, Music, Politics, and Science, enabling systematic evaluation of model generalization across diverse domains.

\subsection{Evaluation Metrics} We use micro-F1, a widely adopted metric in Named Entity Recognition, as the evaluation measure.

\subsection{Compared Baselines} We select the following representative zero-shot and out-of-domain Named Entity Recognition methods as baselines for comparison:
\begin{itemize}
    \item \textbf{InstructUIE} \cite{instructuie} InstructUIE jointly instruction tuning multiple IE tasks under a unified framework.
    \item \textbf{UniNER} \cite{universalNER}  UniNER combines target distillation with instruction tuning for cross-task generalization.
    \item \textbf{IEPile} \cite{iepile} This method constructs a large data set to train the information extraction ability of the model. 
    \item \textbf{GoLLIE} \cite{GOLLIE} GoLLIE trains LLMs with structured annotation guidelines for type-driven extraction.
    \item \textbf{KnowCoder} \cite{knowcoder} KnowCoder encodes structured constraints as executable code to enhance semantic understanding.
    \item \textbf{GLiNER} \cite{gliner} a lightweight general-purpose named entity recognition model based on bidirectional Transformer, which performs exceptionally well in resource-constrained scenarios.
    \item \textbf{IRRA} \cite{IRRA} IRRA applies retrieval augmented generation to refine entity typing. Since the optimal settings introduce additional knowledge, but the Guidelines settings are similar to ours, we chose to use the Guidelines settings for comparison.
    \item \textbf{GUIDEX} \cite{guidex} GUIDEX performs data synthesis guided by the schema and infers entities using type definitions.
\end{itemize}

For all baselines, we report the best results under the zero-shot and out-of-domain settings as documented in the original papers.

\subsection{Backbones \& Implementation} In all experiments, we consistently used Llama3-8B-Instruct as the backbone model. Additionally, to validate the generalization ability of the proposed method across different model families, we further introduced Qwen2.5-7B-Instruct as a supplementary backbone model. It should be noted that since the Anchor Extractor utilizes the Llama3-8B LoRA weights provided by IEPile\cite{iepile}, Qwen2.5-7B-Instruct is used only as the base model for the Explorer Extractor, Archetype Classifier, and Type Calibrator in this setup. Both the extractors and the classifier are trained via supervised instruction tuning, with fine-tuning that updates only a small subset of parameters using LoRA \cite{lora}. All models are trained on three NVIDIA RTX 3090 GPU using the LlamaFactory framework \cite{llamafactory}.

\section{Results}
Table ~\ref{tab:result} presents a comparison between our SAM-NER and several state-of-the-art zero-shot NER systems. When using Llama3-8B as the backbone, our method achieved an average F1 score of 66.3 on the CrossNER, outperforming all comparison methods in the benchmark. Specifically, SAM-NER achieved the best performance in the literature, music, and science domains, with a 4.3-point increase in F1 score over the second-best method in the literature domain. When using Qwen2.5-7B as the backbone, the model’s average F1 score also approached the runner-up’s level, further demonstrating the effectiveness of the proposed abstract mediation mechanism in stabilizing cross-domain transfer.

We observe relatively lower performance in the AI; However, UniNER and IEPile-Llama3-8B, trained on the same data as our Cooperative extractor, exhibit similar trends, suggesting that this problem may stem from an intrinsic bias in the training data. GLiNER's performance advantages in the political domain may stem from the stability of its entity structure and clear semantic boundaries, characteristics that align highly effectively with its entity span-type semantic matching mechanism. When the semantic distance between target domain types and the abstract mediation layer is insufficient, the prototype's mediating role may become less effective.

\section{Analysis}
\label{sec:analysis}
\paragraph{Contribution of Different Components.} To demonstrate the contributions of the  entity discovery via cooperative extraction and definition-guided semantic calibration stages to SAM-NER under the ZS-NER setting, we conducted further analysis. During the entity discovery via cooperative extraction phase, we removed the anchor extractor (\textbf{w/o anc.}) and the explorer extractor (\textbf{w/o exp.}) respectively. We removed the definition-guided semantic calibration stage (\textbf{w/o cali.}) and retrained the classification model on data with unmapped abstract Archetypes to adapt it for predicting target entity types.

\begin{table}[h]
    \centering
    \small
    \setlength{\tabcolsep}{6pt}
    \begin{tabular}{c|ccc}
    \toprule
        \textbf{Dataset} & \textbf{w/o exp.} & \textbf{w/o anc.} & \textbf{w/o cali.} \\
    \midrule
        AI & \(\text{53.0}_{\text{-5.2}}\) & \(\text{54.1}_{\text{-4.1}}\) & \(\text{48.5}_{\text{-9.7}}\) \\
        Literature & \(\text{64.6}_{\text{-4.1}}\) & \(\text{61.9}_{\text{-6.8}}\) & \(\text{56.1}_{\text{-12.6}}\) \\
        Music & \(\text{65.3}_{\text{-5.9}}\) & \(\text{66.1}_{\text{-5.1}}\) & \(\text{58.6}_{\text{-12.6}}\) \\
        Politics & \(\text{63.6}_{\text{-4.6}}\) & \(\text{61.9}_{\text{-6.3}}\) & \(\text{63.7}_{\text{-4.5}}\) \\
        Science & \(\text{61.4}_{\text{-3.7}}\) & \(\text{58.8}_{\text{-6.3}}\) & \(\text{54.1}_{\text{-11}}\) \\
    \bottomrule
    \end{tabular}
    \caption{Micro-F1 Values for Different Components in SAM-NER}
    \label{tab: component removal}
\end{table}

The results are shown in Table~\ref{tab: component removal}. Can be observed that:
(1) Removing either extractor (w/o anc. or w/o exp.) consistently leads to performance degradation. This is mainly because the two extractors are trained on different data distributions and supervision signals, resulting in divergent entity boundary predictions.
(2) The performance drops substantially under the w/o cali. setting, which highlights the critical role of the intermediate semantics in out-of-domain generalization. Due to the mismatch between the intrinsic semantic representations of the language model and the target-domain type definitions, the model struggles to directly interpret target type descriptions—especially for novel types that never appear in the training data—thereby causing classification errors.

\begin{table}[h]
    \centering
    \small
    \setlength{\tabcolsep}{0pt} 
    \begin{tabular*}{\columnwidth}{@{\extracolsep{\fill}} l ccccc}
    \toprule
        \textbf{Method} & \textbf{AI} & \textbf{Literature} & \textbf{Music} & \textbf{Politics} & \textbf{Science}\\
    \midrule
        w/o CCR & 50.8 & 65.3 & 67.2 & 65.5 & 60.9 \\
        w/ CCR & \textbf{58.2} & \textbf{68.7} & \textbf{71.2} & \textbf{68.2} & \textbf{65.1} \\
    \bottomrule
    \end{tabular*}
    \caption{Impact of the Collaborative Consensus Refinement(CCR) strategy on performance.}
    \label{tab:CCR removal}
\end{table}

\begin{figure}[t]
  \includegraphics[width=1 \columnwidth]{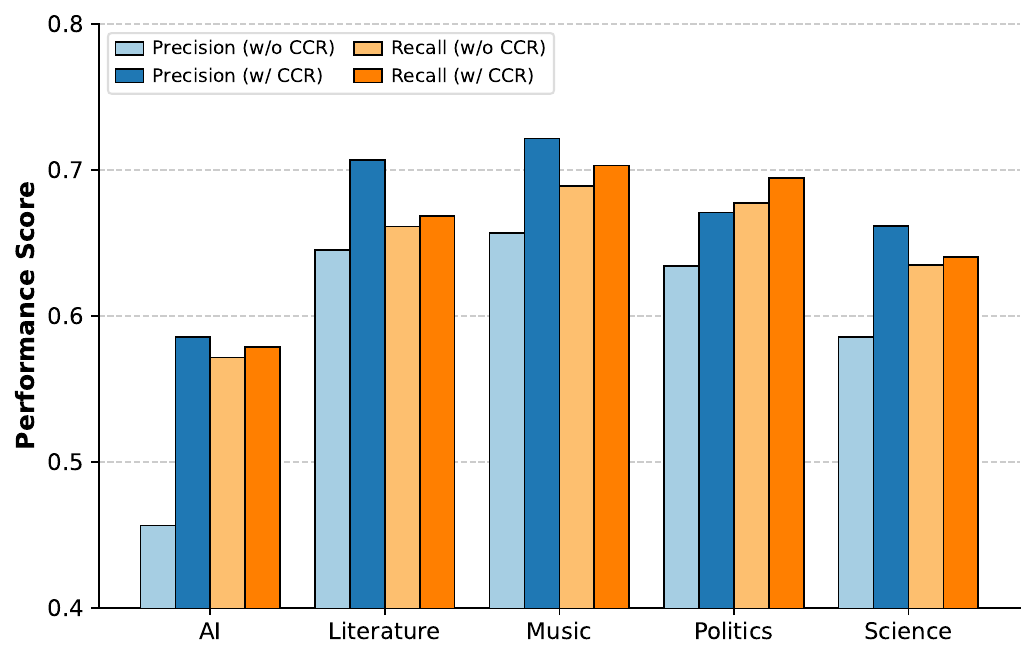}
  \caption{Impact of CCR strategy on precision and Recall in different domains}
  \label{fig:CCR P/R}
\end{figure}

\paragraph{Contribution of Collaborative Consensus Refinement.}
Drawing on the results in Table ~\ref{tab:CCR removal} and Figure ~\ref{fig:CCR P/R}, the Collaborative Consensus Refinement strategy effectively mitigates the spurious mention activations typical of explorer extractor trained on silver-standard data, which often misidentify generic functional terms as entity mentions. As illustrated in Table ~\ref{tab:CCR removal}, the integration of CCR yields consistent F1 gains across all domains, highlighted by a 7.4-point increase in the AI domain. Metrics in Figure ~\ref{fig:CCR P/R} further confirm that this improvement stems from a significant optimization in precision, successfully rectifying the precision-recall imbalance of the base extractor. By leveraging dual-model consensus as a semantic filter, CCR prunes generic lexical noise while preserving the coverage of long-tail entities. Our collaborative mechanism effectively reconciles the extensive discovery capabilities of silver-data models with the high-precision reasoning of models trained on high-quality instructions.

\paragraph{Why 14 Semantic Archetypes?}
To derive a universal abstract mediation space, we performed semantic clustering on entity patterns distilled from the IEPile corpus. In determining the optimal cluster number $k$, we adhered to the principle of ontological parsimony, seeking a balance between semantic granularity and cross-domain stability.

\begin{figure}[h]
  \includegraphics[width=1 \linewidth]{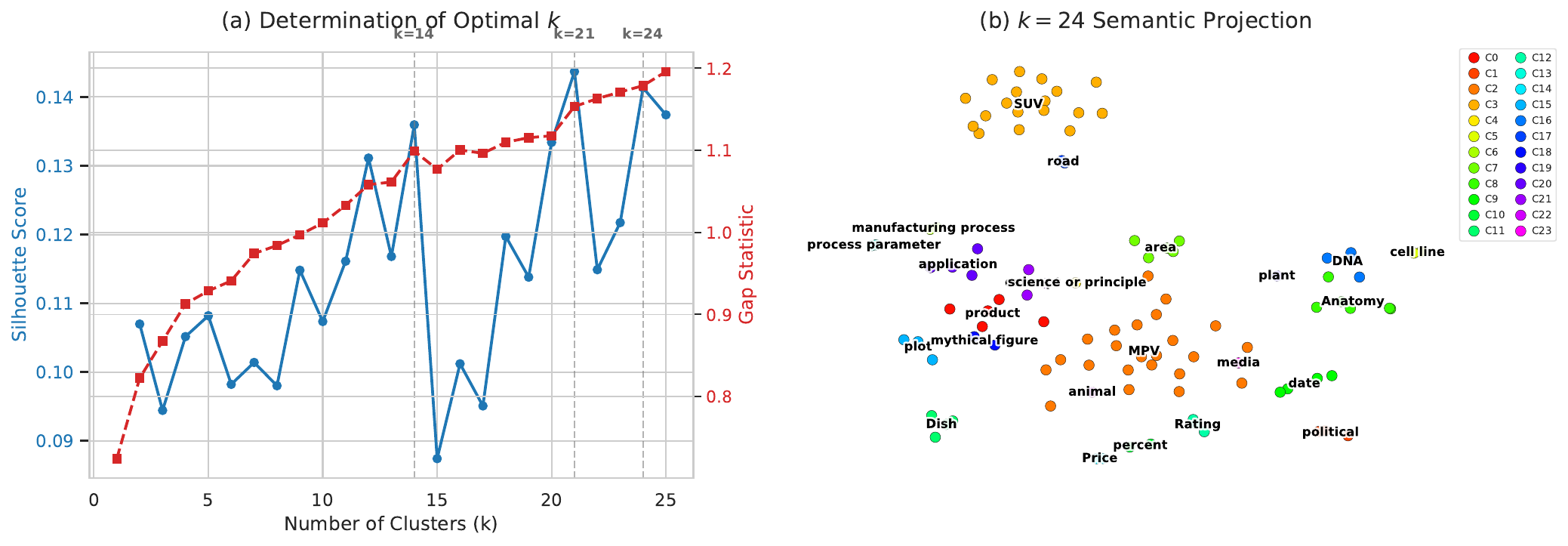}
  \caption{The Cluster Analysis. The left figure (a) shows the scores of Silhouette and Gap at different k values. The right figure (b) presents the clustering results when $k=24$.}
  \label{fig:cluster}
\end{figure}

As illustrated in Figure ~\ref{fig:cluster}(a), the Silhouette Score achieves a prominent local peak at $k=14$. While higher numerical scores appear at $k=21$ and $k=24$, the Gap Statistic reveals that the structural gains from increasing k exhibit diminishing marginal returns beyond 14. Furthermore, the observed drop in the Gap Statistic at $k=15$ signals a potential instability in semantic boundaries at that specific dimensionality. Although Figure ~\ref{fig:cluster}(b) demonstrates that $k=24$ can capture hyper-specific semantic nuances (e.g., "SUV" or "MPV"), such high-specificity clusters are prone to coupling with domain-specific noise in zero-shot scenarios, thereby compromising the transferability of the mediation space. Consequently, we set $k=14$ as it represents the optimal trade-off between intra-cluster cohesion and inter-cluster separability, establishing a robust semantic space that effectively resists domain shift.

\section{Conclusion}
In this paper, we presented \textbf{SAM-NER}, a three-stage framework that mitigates semantic drift in zero-shot NER through \emph{Semantic Archetype Mediation}. By decoupling span discovery from fine-grained target typing, SAM-NER stabilizes cross-domain transfer under heterogeneous label semantics. Concretely, we first perform \emph{Entity Discovery} via cooperative extraction and consensus-based denoising to obtain a high-fidelity candidate set, then conduct \emph{Abstract Mediation} by projecting candidates into a universal semantic archetype space to provide domain-invariant anchors, and finally apply \emph{Definition-Guided Calibration} to ground archetype-level predictions into target-domain types through constrained, definition-aligned inference. Experiments on CrossNER demonstrate that SAM-NER consistently outperforms strong prior zero-shot NER baselines in cross-domain settings, with particularly pronounced gains in domains where schema-specific label semantics amplify misalignment. Overall, our findings highlight semantic archetype mediation as an effective and interpretable strategy for improving domain-invariant generalization in zero-shot NER.

\section*{Limitations}
Despite its strong empirical performance, \textbf{SAM-NER} has two main limitations that warrant further investigation: \textbf{(1) Taxonomic Bias and Bounded Universality.}
SAM-NER relies on a compact set of 14 semantic archetypes distilled from IEPile, which does not constitute a theoretically exhaustive ontology. As a result, the universality of the archetype space is bounded by the coverage and granularity of the distillation source. In highly specialized domains, limited semantic resolution may lead to \emph{coarse-mapping} errors, where domain-specific nuances are absorbed into overly broad archetypes, reducing precision under fine-grained or atypical target taxonomies. \textbf{(2) Dependence on Definition Discriminability.}
The fidelity of definition-guided calibration depends on the linguistic discriminability of target type definitions. When definitions are underspecified, highly overlapping, or inconsistent across labels, the constrained alignment process becomes sensitive to description quality and may produce unstable type assignments. Therefore, while semantic archetype mediation mitigates semantic drift at the schema level, the final predictions remain partly contingent on the clarity and separability of human-authored label definitions.

\section*{Acknowledgments}
This research was supported in part by National Science and Technology Major Project (2021ZD0111502), Natural Science Foundation of China (U24A20233, 62406078, 62476163), the Guangdong Basic and Applied Basic Research Foundation (2023B1515120020), CCF-DiDi GAIA Collaborative Research Funds (CCF-DiDi GAIA 202521), and Guangdong Laboratory of Artificial Intelligence and Digital Economy (SZ)(GML-KF-24-23).

\bibliography{custom}

\clearpage

\section*{}
\appendix
\section{Prompts}
\label{app:appendixA}

\begin{figure*}[h]
\centering
\includegraphics[width=1\linewidth]
{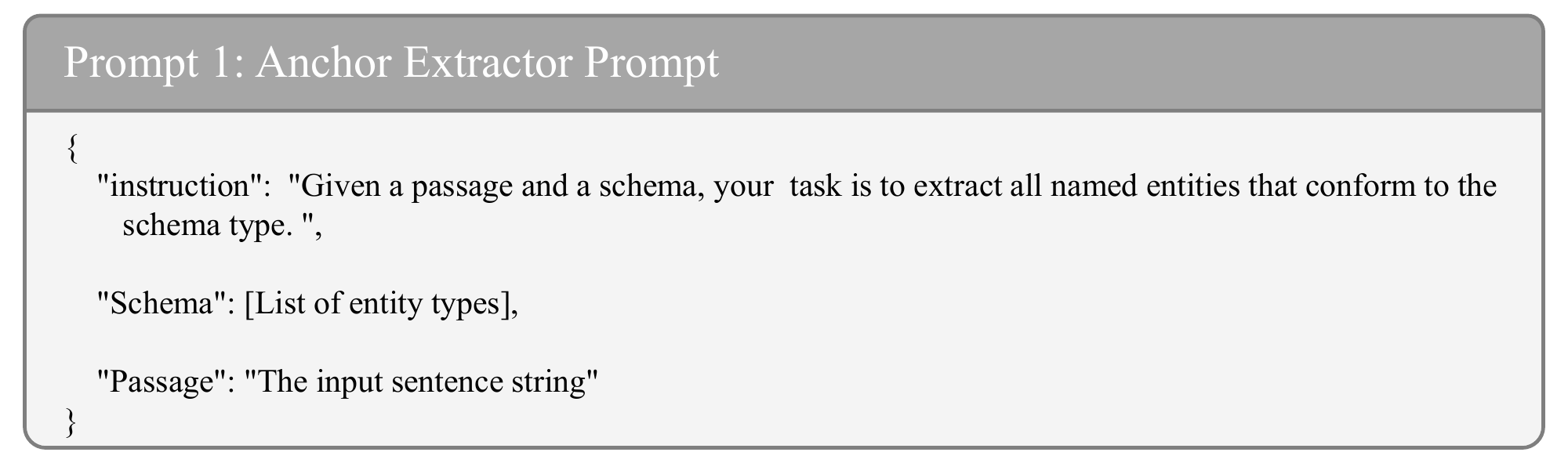}
  \caption{Anchor Extractor Prompt. Guiding the model to identify and extract entities from text following predefined extraction schema.}
  \label{fig:prompt1}
\end{figure*}

\begin{figure*}[h]
\centering
  \includegraphics[width=1\linewidth]{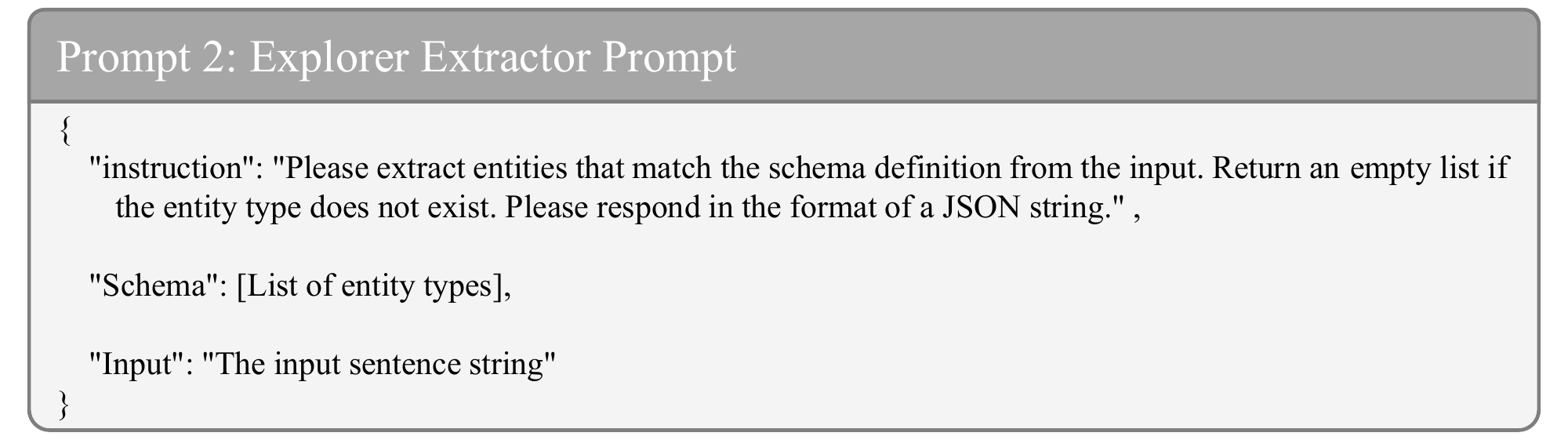}
  \caption{Explorer Extractor Prompt. Guiding the model to identify and extract entities from text following predefined extraction schema.}
  \label{fig:prompt2}
\end{figure*}

\begin{figure*}[h]
  \includegraphics[width=1\linewidth]{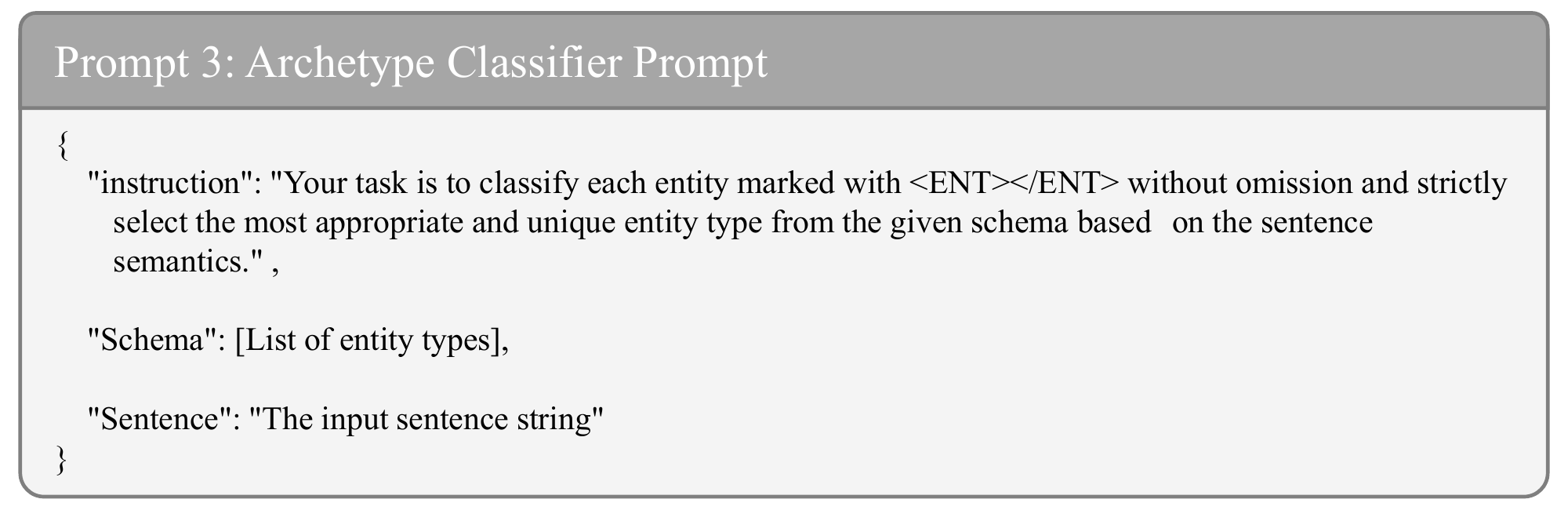}
  \caption{Archetype Classifier Prompt. Guiding the model to assign semantically broad types to annotated entities in the input text based on predefined abstract schema.}
  \label{fig:prompt3}
\end{figure*}

\begin{figure*}[h]
  \includegraphics[width=1\linewidth]{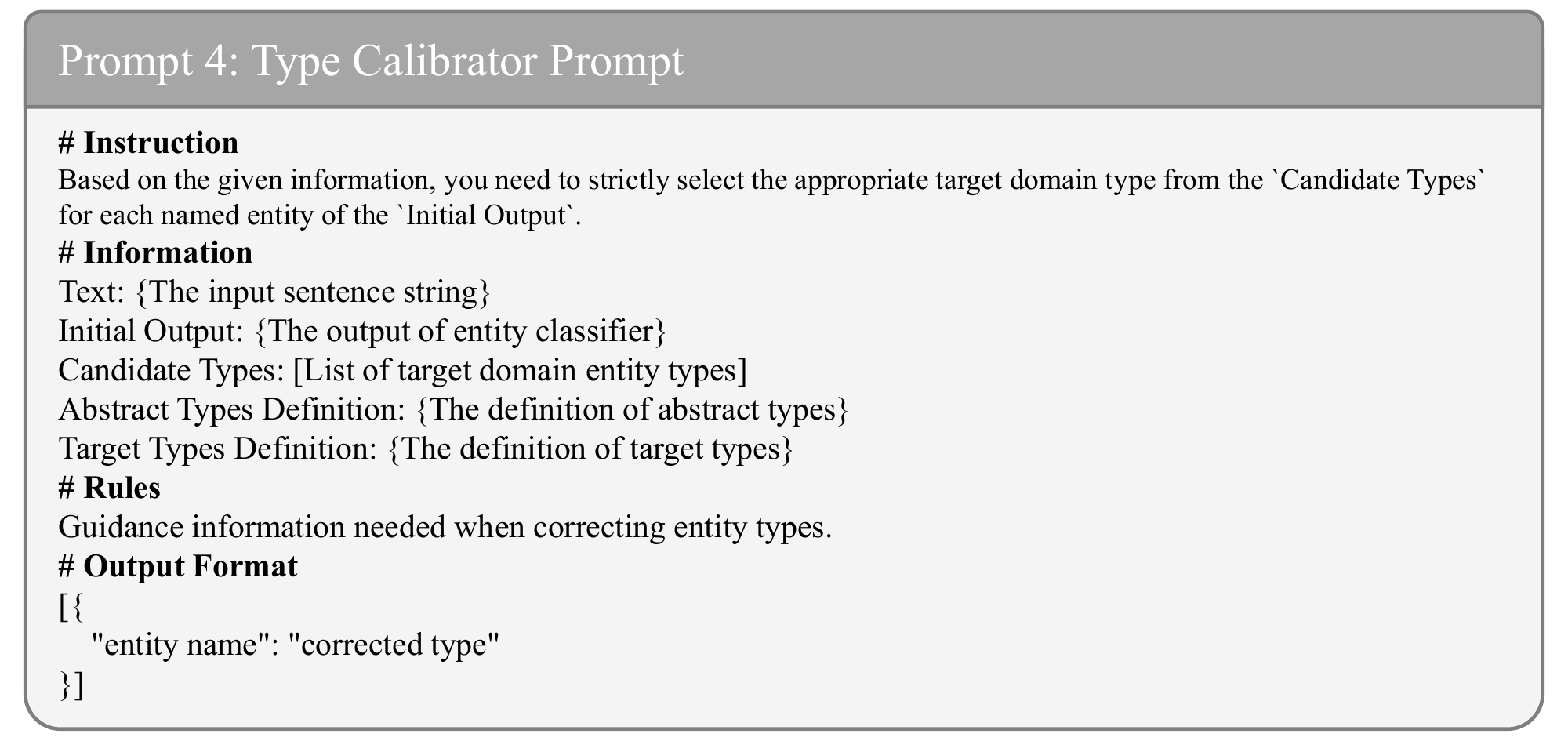}
  \caption{Type Calibrator Prompt. Guiding the model to calibrate the abstract types assigned to entities to target-domain types based on both semantically broad type definitions and target-domain semantic type definitions.}
  \label{fig:prompt4}
\end{figure*}

The full set of prompts used in SAM-NER is illustrated in Figures~\ref{fig:prompt1}--\ref{fig:prompt4}.
These prompts decompose the NER process into four progressive stages: anchor extraction, entity exploration, archetype classification, and type calibration.

\section{Detailed mapping relationship in IEPile and definition of archetype}
\label{app:appendixC}

Following the empirical determination of \textit{k=14}, we conducted a semantic consolidation of labels from the IEPile corpus to derive 14 abstract types. The systematic projection from fine-grained entity categories to these universal archetypes is summarized in Table ~\ref{tab:abstract_type_mapping}. And the detailed definition of the prototype is shown in Table ~\ref{tab:archetype_definition}

\renewcommand{\tabularxcolumn}[1]{m{#1}}

\begin{table*}[t]
\centering
\small
\renewcommand{\arraystretch}{1.5}
\setlength{\tabcolsep}{10pt}
\begin{tabularx}{\textwidth}{>{\bfseries\RaggedRight}m{3cm} >{\RaggedRight\arraybackslash}X} 
\toprule
\textbf{Abstract Type} & \textbf{Fine-grained Entity Types} \\ 
\midrule
    Person & actor, character, director, mythical figure, person \\
    Organization & organization, media \\
    Location & Amenity, Location, location, exact location, geographical phenomenon, geographical social political, facility, road, river, area \\
    Biology & animal, plant, biology \\
    Medicine & Anatomy, DNA, RNA, GENE, protein, cell line, cell type, disease, Disease, biomedical, medicine \\
    Food & Cuisine, Dish, food, Restaurant Name, review \\
    Vehicle & vehicle, vehicle model, vehicle range, vehicle type, vehicle velocity, brand of vehicle, color of vehicle, orientation of vehicle, position of vehicle, estate car, SUV, MPV, hatchback, roadster, sports car, sedan, coupe, trailer, van, truck, motorcycle, vintage car, bus \\
    Creative\_Work & song, work of art, title, movie, genre, creative\_work \\
    Event & event, plot \\
    Artifact & instrument, product, artifact \\
    Computer\_Science & application, enabling technology, concept or principle, process characterization, process parameter, machine or equipment, engineering feature, machanical property, manufacturing process, manufacturing standard, computer\_science \\
    Political & law, national religious political, political \\
    Science & astronomical object, language, material, Chemical, science  \\
    Misc & misc, else \\
\bottomrule
\end{tabularx}
\caption{The detailed mapping relationship between abstract type and fine-grained entity type in IEPile.}
\label{tab:abstract_type_mapping}
\end{table*}

\begin{table*}[t]
\centering
\small
\renewcommand{\arraystretch}{1.5}
\setlength{\tabcolsep}{10pt}
\begin{tabularx}{\textwidth}{>{\bfseries\RaggedRight}m{3cm} >{\RaggedRight\arraybackslash}X} 
\toprule
\textbf{Archetype} & \textbf{Archetype Definition} \\ 
\midrule
    Person & Entities representing specific individuals identified by their proper names, including real people, fictional characters, nicknames, and aliases. \\
    Organization & Entities representing structured groups of people working together for a common purpose, including corporations, government agencies, NGOs, musical bands, political parties, and educational institutions. \\
    Location & Entities representing spatial or geographic regions, including countries, cities, administrative divisions, physical facilities, landmarks, and public spaces. \\
    Biology & Entities related to living organisms and taxonomy, including animal and plant species, families, and general biological classifications. \\
    Medicine & Entities related to healthcare and biomedical sciences, including diseases, drugs, medical procedures, anatomical structures, physiological processes, and clinical concepts. \\
    Food & Entities related to consumables, including ingredients, prepared dishes, beverages, and culinary concepts. \\
    Vehicle & Entities representing manufactured devices designed for transportation, including cars, aircraft, ships, spacecraft, and their specific models or classes. \\
    Creative\_Work & Entities representing distinct artistic or intellectual creations, such as books, songs, movies, video games, software titles, and media franchises. \\
    Event & Entities representing specific occurrences or organized activities happening at a specific time and place, including festivals, wars, sports matches, conferences, and natural disasters. \\
    Artifact & Entities representing man-made objects with specific functions, including tools, instruments, gadgets, weapons, and consumer goods. \\
    Computer\_Science & Entities related to computing and technology, including programming languages, algorithms, software architectures, technical protocols, digital metrics, and IT terminology. \\
    Political & Entities related to governance, social structures, and ideologies, including laws, treaties, policies, religious groups, ethnicities, and sociopolitical systems. \\
    Science & Entities related to scientific disciplines and natural phenomena, including academic fields, chemical elements, compounds, celestial bodies, and scientific theories.  \\
    Misc & Entities that cannot be clearly classified into the specific categories above, serving as a catch-all for other named entities. \\
\bottomrule
\end{tabularx}
\caption{The detailed definition of archetype.}
\label{tab:archetype_definition}
\end{table*}

\section{Complexity Analysis of SAM-NER}
\label{app:appendixD}

To evaluate the efficiency of SAM-NER, we reported the time cost and memory consumption for different module combinations. Note that all experiments were conducted on the NVIDIA A800 GPU. To objectively reflect SAM-NER’s raw computational overhead, we used the Llama-Factory framework’s default inference pipeline without FlashAttention, quantization, or any third-party acceleration engines, in order to provide a baseline performance reference. The results are presented in Table ~\ref{tab:complexity_analysis}.

\begin{table*}[h]
\centering
\small
\renewcommand{\arraystretch}{1.5}
\setlength{\tabcolsep}{18pt}
\begin{tabular}{l c c c}
\toprule
\textbf{Method} & \textbf{Time Cost} & \textbf{Memory Consumption} & \textbf{Avg. score} \\ 
\midrule
    w/o exp.   & 6486 & 1GPU $\times$ 29.53GB & 61.6 \\
    w/o anc.   & 6507 & 1GPU $\times$ 29.53GB & 60.6 \\
    w/o cali.  & 2883 & 1GPU $\times$ 17.78GB & 56.2 \\
    w/ all     & 7247 & 1GPU $\times$ 29.53GB & 66.3 \\
\bottomrule
\end{tabular}
\caption{Complexity Analysis of SAM-NER on CrossNER. The time unit is seconds. "w/o exp." refers to without employing explorer extractor."w/o anc." refers to without using anchor extractor."w/o cali." refers to without employing definition-guided semantic calibration stage. "w/ all" refers to using all components and stages.}
\label{tab:complexity_analysis}
\end{table*}

\section{Hyperparameter Settings for SAM-NER}
\label{app:appendixE}

The hyperparameters of SAM-NER are presented in Table~\ref{tab:Hyperparameter}. 

\begin{table*}[t]
\centering
\small
\renewcommand{\arraystretch}{1.5}
\setlength{\tabcolsep}{18pt}
\begin{tabular}{l|c|c}
\toprule
\textbf{Hyperparameter} & \textbf{Explorer Extractor} & \textbf{Archetype Classifier} \\ 
\midrule
    loRA\_rank(r) & 8 & 8 \\
    loRA\_alpha($\alpha$) & 16 & 16 \\
    cutoff\_len & 1024 & 1024 \\
    per\_device\_train\_batch\_size & 2 & 2 \\
    gradient\_accumulation\_steps & 8 & 8 \\
    learning\_rate & 2.0e-5 & 3.0e-5 \\
    num\_train\_epochs & 3.0 & 3.0 \\
    lr\_scheduler\_type & cosine & cosine \\
    warmup\_ratio & 0.05 & 0.05 \\
    dtype & bf16 & bf16 \\
\bottomrule
\end{tabular}
\caption{Hyperparameter Settings of SAM-NER}
\label{tab:Hyperparameter}
\end{table*}

\end{document}